\documentclass{ieeetj}
\usepackage{cite}
\usepackage{amsmath,amssymb,amsfonts}
\usepackage{algorithmic}
\usepackage{graphicx,color}
\usepackage{textcomp}
\usepackage{xcolor}
\usepackage{hyperref}
\hypersetup{hidelinks=true}
\usepackage{algorithm,algorithmic}
\usepackage{booktabs}
\usepackage{multirow}
\usepackage{pgfplots}

\def\BibTeX{{\rm B\kern-.05em{\sc i\kern-.025em b}\kern-.08em
    T\kern-.1667em\lower.7ex\hbox{E}\kern-.125emX}}
\AtBeginDocument{\definecolor{tmlcncolor}{cmyk}{0.93,0.59,0.15,0.02}\definecolor{NavyBlue}{RGB}{0,86,125}}

\def\OJlogo{\vspace{-4pt}$<$Society logo(s) and publication title will appear here.$>$}
\def\seclogo{\vspace{10pt}$<$Society logo(s) and publication title will appear here.$>$}

\def\authorrefmark#1{\ensuremath{^{\textbf{#1}}}}

\begin{document}
\receiveddate{XX Month, XXXX}
\reviseddate{XX Month, XXXX}
\accepteddate{XX Month, XXXX}
\publisheddate{XX Month, XXXX}
\currentdate{XX Month, XXXX}
\doiinfo{TBD}

\markboth{}{Sheppard {et al.}}

\title{MarsLGPR: Mars Rover Localization with Ground Penetrating Radar}

\author{Anja Sheppard\authorrefmark{1}, Student Member, IEEE, Katherine A. Skinner\authorrefmark{1}, Member, IEEE}
\affil{Department of Robotics, University of Michigan, Ann Arbor, MI 48109 USA}
\corresp{Corresponding author: Anja Sheppard (email: anjashep@umich.edu).}
\authornote{This work was supported in part by NSF Grant \#DGE 2241144, the University of Michigan Robotics Department, and the University of Michigan Space Institute Pathfinder Grant. This work is also supported in part by NASA Pennsylvania Space Grant Consortium (PSGC) Mini-Grants Program and Duquesne BME departmental funds. \\ \\This work has been submitted to the IEEE for possible publication. Copyright may be transferred without notice, after which this version may no longer be accessible.\vspace{-3em}}

\begin{abstract}
In this work, we propose the use of Ground Penetrating Radar (GPR) for rover localization on Mars. Precise pose estimation is an important task for mobile robots exploring planetary surfaces, as they operate in GPS-denied environments. Although visual odometry provides accurate localization, it is computationally expensive and can fail in dim or high-contrast lighting. Wheel encoders can also provide odometry estimation, but are prone to slipping on the sandy terrain encountered on Mars. Although traditionally a scientific surveying sensor, GPR has been used on Earth for terrain classification and localization through subsurface feature matching. The Perseverance rover and the upcoming ExoMars rover have GPR sensors already equipped to aid in the search of water and mineral resources. We propose to leverage GPR to aid in Mars rover localization. Specifically, we develop a novel GPR-based deep learning model that predicts 1D relative pose translation. We fuse our GPR pose prediction method with inertial and wheel encoder data in a filtering framework to output rover localization. We perform experiments in a Mars analog environment and demonstrate that our GPR-based displacement predictions both outperform wheel encoders and improve multi-modal filtering estimates in high-slip environments. Lastly, we present the first dataset aimed at GPR-based localization in Mars analog environments, which will be made publicly available at \url{https://umfieldrobotics.github.io/marslgpr/}.
\end{abstract}

\begin{IEEEkeywords}
ground penetrating radar, localization, multi-modal perception, space robotics
\end{IEEEkeywords}


\maketitle

\vspace{-1em}

\section{INTRODUCTION}

\IEEEPARstart{P}{recisely} localizing planetary rovers in GPS-denied environments such as the Moon and Mars is a challenging task. Until recently, localization for the Perseverance rover on the surface of Mars has required a human-in-the-loop for manual corrections and tuning \cite{li2006, verma2024}. New approaches have seen increasingly autonomous solutions for global state estimation \cite{nash2024}, but this problem has not been entirely solved. As interest increases in launching lower-cost rovers with shorter lifespans, there is a need for real-time GPS-denied rover localization that does not require a large team of human operators in the loop.

{Wheel encoder and inertial sensor data are commonly used to localize a robot. However, wheel encoders are notoriously prone to incorrect predictions due to wheel slip. The surface of Mars has many sandy areas that have caused drastic enough wheel slip and sinkage to permanently trap the\unskip\parfillskip 0pt \par}

\begin{figure}[H]
    \includegraphics[width=0.95\linewidth]{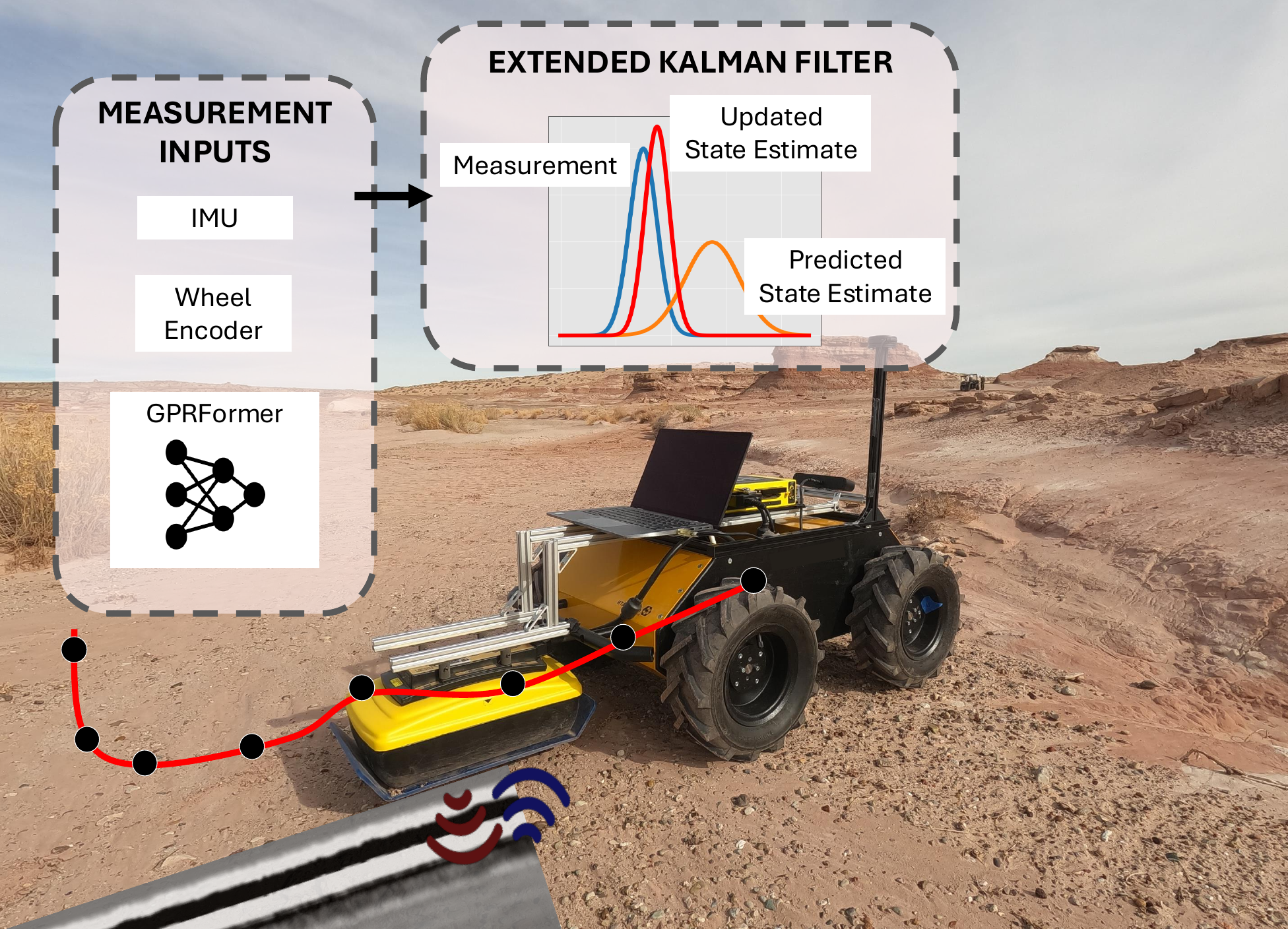}
    \caption{Our mobile robot collection platform in a Mars analog environment. The proposed method demonstrates a novel approach to relative displacement estimation with Ground Penetrating Radar, which then informs a filtering-based state estimation algorithm.}
    \label{fig:cover}
\end{figure}

\begin{table*}[!ht]
    \centering
    \caption{Localizing Ground Penetrating Radar datasets.}
    \begin{tabular}{lcccccc}
        \toprule
        Dataset & RGB-D & GPR & IMU & Ground Truth & Location & Publicly Available \\
        \midrule
        \midrule
        GROUNDED \cite{ort2023} & FLIR Grasshopper & Custom LGPR & RT3003 IMU & OXTS RT3003 GPS & Highway & No \\
        & Velodyne HDL-64 & & & & Rural Road \\
        & & & & & Urban Road \\
        \midrule
        CMU-GPR \cite{baikovitz2021cmugpr} & Realsense D435 & NOGGIN 500 & XSENS MTi-30 & Leica TS15 Station GPS & Basement & Yes\\
        & & & & & Factory Floor \\
        & & & & & Parking Garage \\
        \midrule
        MarsLGPR (Ours) & ZED2i & NOGGIN 500 & ZED2i & RTAB-Map & Mars Analog & Yes \\
        \bottomrule
    \end{tabular}
    \label{tab:datasets}
\end{table*}

\noindent Spirit rover \cite{townsend2014}, making encoders an unreliable sensor for localization. Visual odometry (VO) has shown the benefits of including additional sensing modalities to improve state estimation. In VO, features are tracked between subsequent pairs of images in order to characterize the robot's motion and look for loop closure. Although VO typically outperforms inertial and wheel encoder odometry, this method is more computationally costly and is not realistic for rovers to run in real-time given current hardware constraints. In a VO demonstration for Martian rovers, Spirit and Opportunity needed three minutes to complete a single vision-based tracking step \cite{maimone2007}. The Curiosity rover has demonstrated on-board VO tests, but with each image processing step taking around 34 seconds \cite{maimone2022visual}. For reference, state-of-the-art terrestrial VO runs at 30-40 frames per second \cite{campos2021orb}. The Perseverance rover localization framework \cite{nash2024} includes offline optimization-based localization methods, such as matching against existing orbital elevation maps, but does not mention use of real-time VO on board the rover.


In the space domain, scientific sensors can be creatively repurposed to provide information about a robot's environment to the navigation algorithms, such as terrain type or slip characterization \cite{kraut2010, cunningham2019}. Ground Penetrating Radar (GPR) has already demonstrated promise on Earth as a tool for localization by using subsurface features \cite{ort2020, baikovitz2021} and for terrain classification using deep learning methods \cite{sheppard2024}. The Perseverance rover is equipped with the RIMFAX \cite{hamran2020} GPR sensor, and the planned European Space Agency ExoMars rover will have a GPR to search for in-situ drilling sites \cite{ciarletti2017}. The upcoming CADRE mission to the surface of the Moon plans to have several robots -- each equipped with GPR -- collect data cooperatively \cite{croix2024multi}.

In this work, we propose a novel framework that leverages GPR to improve localization of planetary rovers (Figure~\ref{fig:cover}). Our main contributions are as follows:
\begin{itemize}
    \item We develop a novel deep learning transformer-based model for real-time relative pose estimation with GPR.
    \item We integrate GPR pose estimation into a real-time filtering framework for rover localization.
    \item We collect and present the first GPR-based  localization dataset collected in a Mars analog environment, titled MarsLGPR.
\end{itemize}
We validate our proposed method on our dataset collected in a Mars analog environment, as well as on a publicly available GPR localization dataset. Through experiments, we demonstrate that incorporating GPR can improve localization estimation for mobile robot platforms operating on complex terrain.

\section{RELATED WORK}

\subsection{Localization}

The Perseverance rover has a robust global localization pipeline that has achieved sub-meter accuracy \cite{verma2024}. However, it often involves humans-in-the-loop and specially engineered data pipelines that may be too computationally intensive to run locally. As the accessibility and cost of landing a planetary lander increases, there will be a great need for relatively precise rover localization that can run onboard without human support.

There are two main approaches to localization: filtering-based \cite{jetto1999} and optimization-based \cite{baikovitz2021}. Filtering for robot localization involves nonlinear dynamics, making Extended Kalman Filters (EKFs) and Unscented Kalman Filters (UKFs) popular approaches \cite{moore2016}. Optimization-based approaches such as GTSAM \cite{dellaert2012} estimate a pose within a factor graph framework by minimizing the residual from different sensor inputs. Localization with optimization can lead to higher accuracy, but properly constraining the factors for tractable optimization can be a challenging task in real-time with noisy sensors. In this work we use an EKF framework for localization with Inertial Measurement Unit (IMU), wheel encoder, and GPR data. Although EKFs require careful parameter tuning, they are reliable for real-time use.

\subsection{Ground Penetrating Radar}

GPR is a popular sensor in the fields of civil engineering \cite{lai2018}, archaeology \cite{vaughan1986}, environmental science \cite{knight2001}, and planetary geology \cite{olhoeft1998}. The GPR on-board the Perseverance rover has been used to search for water and to characterize the electromagnetic properties of the Martian subsurface \cite{hamran2020}. The signal returns from a GPR allow for a nonintrusive view of subsurface materials and features. Single GPR returns, or A-scans, can be concatenated sequentially to form a B-scan image that encodes 2D spatial information.

Subsurface features are generally more static than visual features above the surface -- for example, a scene might look completely different to a camera after a fresh coat of snow or during a rainstorm. Additionally, processing GPR traces is less computationally intensive than performing visual feature matching. Relying on GPR features as an additional mode of localization could improve absolute positioning of planetary rovers.

\subsection{Localizing Ground Penetrating Radar}

The use of GPR for localization has been under development for the last decade. Although some work has investigated the use of matching single GPR A-scans \cite{skartados2019}, most approaches take advantage of patterns in sequential GPR traces that are present in concatenated B-scans. Early approaches rely on a GPS prior and a pre-registered map in order to match incoming GPR measurements to an existing database of underground features \cite{stanley2013, cornick2016}. In order to reduce reliance on GPS, subsequent methods \cite{ort2020, ni2022, ort2023, bi2023, zhang2024, bi2024, xu2024, li2024} focus on more robust approaches for GPR feature detection that could aid in matching newly collected data with the pre-registered map of GPR traces. Some approaches utilize signal processing for finding and matching features across B-scans \cite{ort2020, bi2024, xu2024, li2024}, while other works use deep learning for feature extraction \cite{ni2022, ort2023, bi2023, zhang2024}.

The need for a pre-registered map greatly constrains the use-case of these methods, and makes deployment in unfamiliar areas more difficult. Recent localizing GPR work has begun to establish approaches for \textit{relative} pose estimation between B-scan submaps using deep learning \cite{baikovitz2021, wickramanayake2022}. The only work that has fused relative GPR localization with other sensors such as IMU and wheel encoders \cite{baikovitz2021} has not investigated the use of this method on complex terrain representative of Mars surface terrain. Our proposed work is the first to demonstrate the use of transformer deep learning architectures for GPR-based relative pose estimation targeted at rugged off-road environments.

\subsection{Localizing Ground Penetrating Radar Datasets}

There are currently two localizing GPR datasets: GROUNDED \cite{ort2023} and CMU-GPR \cite{baikovitz2021cmugpr}. Both datasets contain GPR, RGB, depth, IMU, and GPS data, as shown in Table \ref{tab:datasets}. The location coverage of both datasets includes roads, parking garages, basements, and factory floors. The GROUNDED dataset is not currently available to the public.

While these datasets are useful for accelerating localizing GPR for autonomous ground vehicles, there is no dataset that is specifically aimed at exploring GPR localization on Mars. We propose MarsLGPR, a localizing GPR dataset collected on Martian analog terrain with additional stereo camera, IMU, and reference pose data.

\section{TECHNICAL APPROACH}

In this section, we review the key technical concepts in our approach to GPR-based localization. We first discuss the GPR sensor model and how we can use it to predict relative displacement. Then we discuss the model architecture for GPRFormer, our model that predicts the relative displacement from GPR traces. Finally, we include details of our EKF implementation that leverages GPRFormer predictions for improved localization.

\subsection{Ground Penetrating Radar Sensor Model}

\begin{figure}
    \centering
    \includegraphics[width=\linewidth]{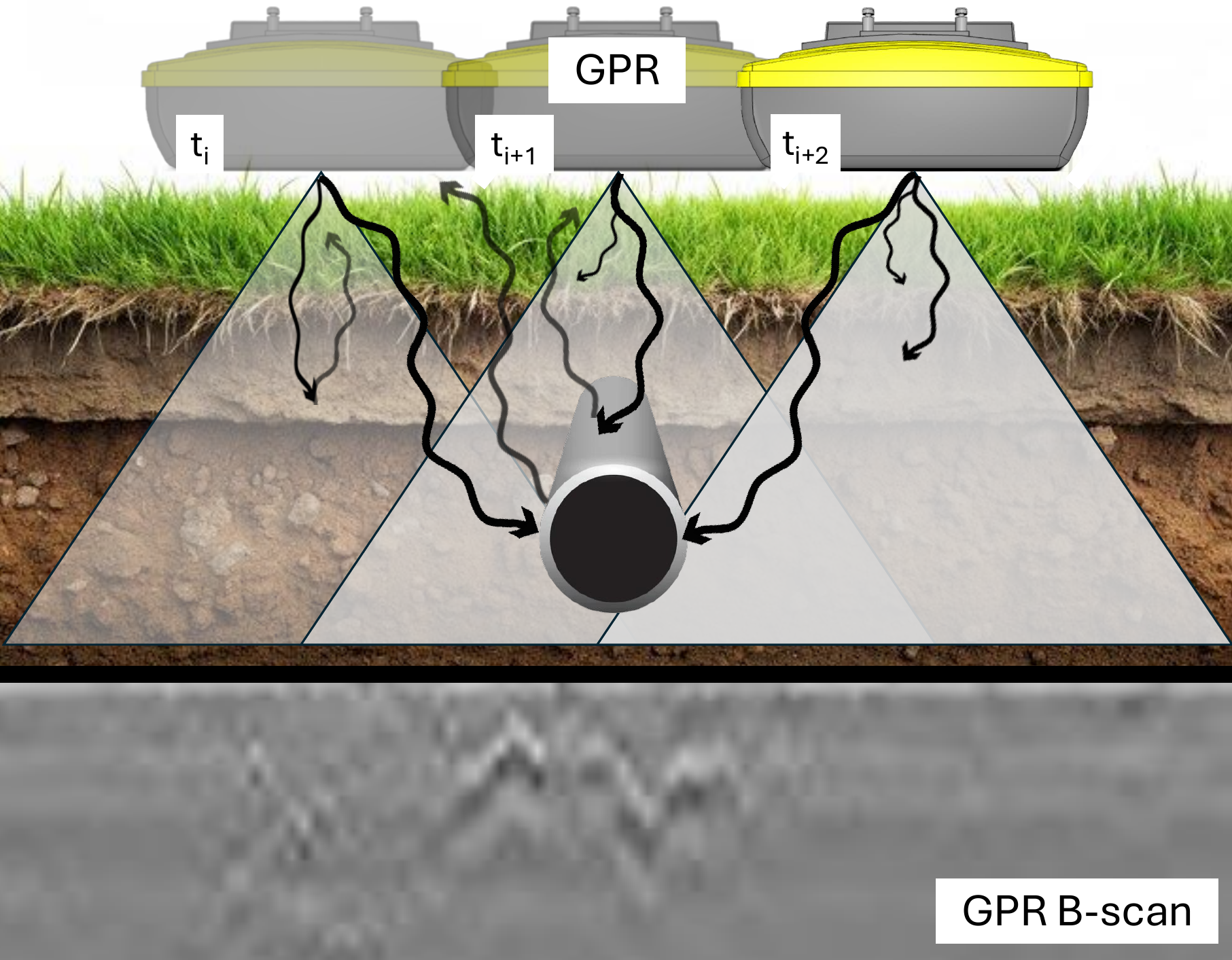}
    \caption{Consecutive GPR samples have overlapping beam spreads, which cause them to detect the same object across multiple returns. Because of this, the resulting GPR B-scan image may not intuitively represent the geometry of objects underground. However, we can use this property of the sensor to predict relative pose translation using deep learning.}
    \label{fig:gpr}
\end{figure}

GPR uses radio waves to characterize subsurface properties, often depicted as a time series B-scan (Figure~\ref{fig:gpr}). Although a B-scan might appear to be a simple pictoral cross-section of the area below the sensor, the sensor returns do not directly capture a visual ``snapshot'' of subterranean feature geometry. This is due to the fact that GPR samples capture relative permittivity between materials \cite{utsi2017}.

When a GPR transmits a radio wave, the reflected energy ($R$) captured in the return is a function of the difference in relative permittivity ($\kappa_1, \kappa_2$) between adjacent materials underground \cite{knight2001}.
\begin{equation}
    R = \frac{\sqrt{\kappa_1} - \sqrt{\kappa_2}}{\sqrt{\kappa_1} + \sqrt{\kappa_2}}
\end{equation}
The larger the difference in the permittivity of the two materials, the stronger the reflected signal will appear. For example, if there is a pipe buried underground, a higher reflectance will be visible at the depth corresponding to the boundary between the top of the object and the surrounding material.

\begin{figure*}
    \centering
    \includegraphics[width=\linewidth]{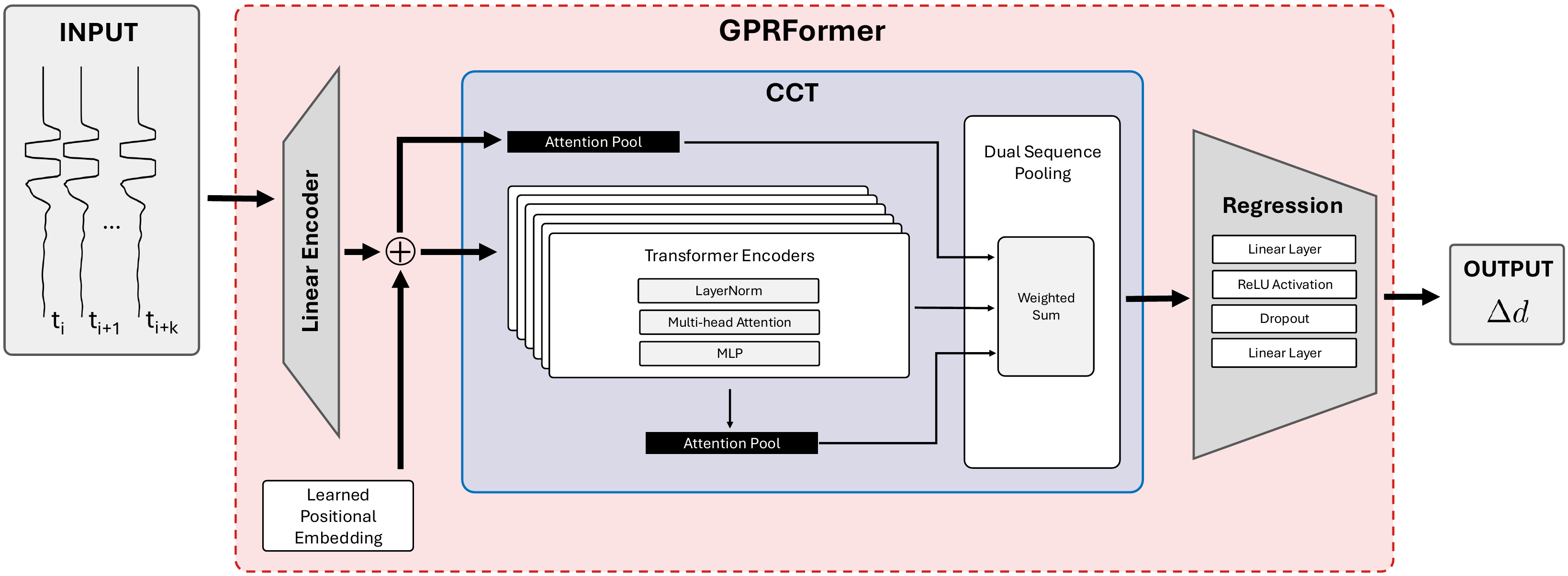}
    \caption{Our GPRFormer network takes in a GPR B-scan with $k$ consecutive traces, performs a linear encoding, adds learned positional embeddings, and then passes the encoded tokens through a compact transformer \cite{hassani2021}. The output from the network is an estimate of relative displacement  ($\Delta d$) that the vehicle traveled while collecting the GPR traces in the input.}
    \label{fig:gprformer}
\end{figure*}

However, the shape of the object visible in the B-scan may not intuitively match the true geometry of the object. A GPR has a wide beam dispersion, so a submerged object will reflect some of the signal back at the transmitter even if the GPR is not directly over the object as shown in Figure~\ref{fig:gpr}. This results in an almost imperceptible correlation of consecutive A-scans. It is this idea that has inspired our investigation into using deep learning for predicting relative pose translation from GPR returns.

Additionally, the frequency of the GPR sensor determines the penetration depth, with a higher frequency corresponding to higher resolution and shallower depth \cite{utsi2017}. The subsurface material properties also impact the penetration depth, with materials such as wet sand or clay reducing signal reach.

\subsubsection{Filtering}
\begin{figure}[t]
    \centering
    \includegraphics[width=\linewidth]{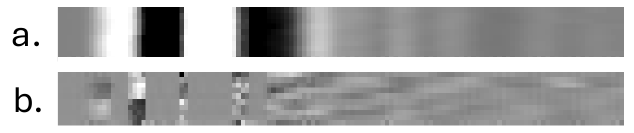}
    \caption{A comparison of a raw B-scan containing $k = 10$ traces (a.) and the filtered result after applying background removal, dewow filtering, SEC gain, and wavelet denoising (b.). Note that the direct wave bands are removed and the deeper features (more rightwards) are more pronounced.}
    \label{fig:filter-comp}
\end{figure}

In order to reduce signal noise and emphasize visible features, signal filtering such as background removal, dewow filters, Spreading and Exponential Compensation (SEC) gain, and wavelet denoising may be utilized \cite{baikovitz2021cmugpr}. An example of pre- and post-filtered B-scan is shown in Figure \ref{fig:filter-comp}.

Background removal is one of the most standard GPR de-noising approaches. It involves removing ``background'' horizontal layering artifacts, typically produced from a direct wave \cite{utsi2017}. We use a global averaging window over the entire batch of traces $\mathbf{G} = [\mathbf{g}_0, \mathbf{g}_1, \dots, \mathbf{g}_{k-1}]$ where $k$ is the number of traces in the network input:
\begin{equation}
    \mathbf{G}_{ij}' = \mathbf{G}_{ij} - \frac{1}{k} \sum_{j=0}^{k-1} \mathbf{G}_{ij}
\end{equation}

Dewow filters remove low-frequency drift from GPR data with polynomial detrending \cite{maruddani2019}. For a given trace $\mathbf{g} \in \mathbb{R}^t$, we find the coefficients $\mathbf{a}$ of a degree-$\rho$ polynomial representing the drift for each trace $\mathbf{g} \in \mathbf{G}$, where $\rho = 3$.
\begin{equation}
    \mathbf{a} = \arg \min \sum_{i=1}^t\left\vert \; \mathbf{g}_i - \sum_{j = 0}^p a_j i^j \; \right\vert^2
\end{equation}
\begin{equation}
    \mathbf{p}_i = \sum_{j=0}^\rho a_j i^j, \;\;\text{ for } i = 0, 1, \dots, t - 1 
\end{equation}
\begin{equation}
    \mathbf{g}' = \mathbf{g} - \mathbf{p}
\end{equation}

SEC gain compensates for the loss of energy as the wavefront propagates through the ground. It does this by scaling the amplitude depending on the order of return \cite{cassidy2008}. Parameter $a$ controls the magnitude of the amplification, while parameter $b$ controls how exponential or linear the amplification is. Threshold $T_\gamma$ sets a maximum amplification at a certain index to avoid over-emphasizing deeper returns (especially when $b$ is closer to 0, and is thus more exponential). We set $a = 0.015$, $b = 0$, and $T_\gamma = 100$.
\begin{equation}
    \boldsymbol{\gamma} = \begin{cases}
        i^b \, e^{a \, i}  & \text{if } i < T_\gamma \\
        {T_\gamma}^b \, e^{a T_\gamma}  & \text{if } i \geq T_\gamma \\
    \end{cases} \;\;\text{ for } i = 0, 1, \dots, t - 1
\end{equation}
$\boldsymbol{\gamma}$ is then multiplied, element-wise, across each row in the B-scan $\mathbf{G}$ \cite{baikovitz2021cmugpr}.

Wavelet denoising aims to separate out the true GPR signal from noise by decomposing the input into the wavelet domain, applying thresholding, and then reconstructing the input in the image space. We selected the Daubechies-6 ``mother" wavelet due to its similarity to a GPR signal and previously-demonstrated superior performance \cite{baili2009}.

In order to reduce signal noise, GPR sensors often collect several samples immediately after one another and average them in a process called ``stacking.'' The stacking process is usually applied during signal postprocessing \cite{utsi2017}. We apply stacking over sets of three traces.

\subsection{GPRFormer}\label{sec:lgpr-network}

Figure \ref{fig:gprformer} shows an overview of our proposed model architecture, GPRFormer, which estimates relative displacement from input GPR data. GPRFormer leverages transformer networks, which uses attention to extract meaningful correlations across tokens or patches in the input data \cite{vaswani2017}. Transformer networks often rely on large datasets, but recent work introduced the Convolutional Compact Transformer (CCT) \cite{hassani2021}, which shows improved performance compared to CNNs on small datasets.

Inspired by methods for 3D relative camera pose estimation \cite{leng2023}, we use a supervised compact transformer architecture to process temporal sequences of 1D sensor data, concatenated into a GPR B-scan. The model consists of three primary components: a linear input encoder that maps 200-dimensional raw sensor readings to 256-dimensional token embeddings, a modified CCT backbone with dual sequence pooling, and a regression head for displacement prediction. Mean Squared Error (MSE) loss is used during training.

\subsubsection{Linear Input Encoder}
The input encoder is a simple linear projection layer that is applied to each of the $k$ consecutive sensor readings to transform them from raw 200-dimensional feature space to a 256-dimensional embedding space. The GPR B-scan format is natively structured and very compact, so we found this simple encoder architecture to be sufficient.

A key component of the model is the positional embedding scheme, which allows the transformer to understand the sequential nature of the input data, as the traces are stacked in order of capture. We specifically utilize learned positional embeddings, which allows the model to adapt positional representations specifically for the sensor data domain. The positional embeddings are added element-wise to the token embeddings before entering the transformer stack.

\subsubsection{CCT Transformer}
Our modified CCT backbone has six transformer encoder layers, each containing pre-normalization, multi-head self-attention with four attention heads, and feed-forward networks each with an expansion ratio of 3.0. The architecture includes a dual sequence pooling mechanism that combines features from both pre-transformer and post-transformer representations. Two separate attention pooling layers generate sequence-wise attention weights, producing pooled representations $\mathbf{x}_1$ (from post-transformer features) and $\mathbf{x}_2$ (from pre-transformer features applied to post-transformer outputs). The final representation combines these via a learnable parameter $\alpha$: 
\begin{equation}
    x_\text{final} = \alpha \mathbf{x}_1 + (1-\alpha) \mathbf{x}_2.
\end{equation}
This allows the model to balance between raw sequential patterns and learned contextual representations from the transformer.

\subsubsection{Regression Head}
The original CCT classification head is replaced with an identity layer, and finally a regression head consisting of two fully connected layers with ReLU activation and dropout ($p=0.1$) predicts the final single displacement distance.

\subsection{Extended Kalman Filter for Localization}

We integrate our predicted GPR displacement in an EKF framework to provide rover localization. EKFs are a popular tool for sensor fusion for robot localization, and they have been well-defined in previous papers \cite{jetto1999, moore2016}. We present an abridged description here to describe our specific implementation.

We make a simplifying assumption that the robot is operating in locally 2D environments in the East North Up (ENU) frame. The sensor mounted on the rear of the robot platform limits its movements to relatively planar regions. We denote the robot's state ($\mathbf{x}_t$) and measurement ($\mathbf{z}_t$) vectors as:
\begin{equation}
    \mathbf{x}_t = \begin{bmatrix} x_t \\ y_t \\ \psi_t \\ \dot{x}_t \\ \dot{y}_t \\ \dot{\psi}_t \\ \ddot{x}_t \\ \ddot{y}_t \end{bmatrix}, \;\;
    \mathbf{z}_t = \begin{bmatrix} x_t^r \\ y_t^r \\ \dot{x}_t^w \\ \dot{\psi}_t^w \\ \psi^m_t \\ \dot{\psi}^g_t \\ \ddot{x}^a \\ \ddot{y}^a \end{bmatrix}
\end{equation}
where $t$ is the continuous time, $(x, y)$ is the position, $\psi$ is the yaw angle, $\cdot$ denotes velocity and $\cdot\cdot$ denotes acceleration. The estimation of a robot's full pose at discrete timestep $k$ can be described with the state equation:
\begin{align}
    \mathbf{x}_k &= f(\mathbf{x}_{k-1}) + w_{k-1},
\end{align}
where $w_{k-1}$ is the process noise. The measurement model represents the measurements we receive from our sensors (wheel encoder, IMU, and GPR) at each timestep:
\begin{align}
    \mathbf{z}_k &= h(\mathbf{x}_k) + v_k.
\end{align}
where $v_k$ is the measurement noise.

We use the EKF implementation in the Robot Operating System (ROS) package $\texttt{robot\_localization}$ \cite{moore2016}.


\subsubsection{Wheel Odometry}

The following differential drive integration method is used to calculate the linear ($\dot{x}^w$) and rotational ($\dot{\psi}^w$) velocity of the vehicle in the robot body frame using the encoder values of all four wheels.
We calculate the left and right wheel velocities by $v_{l/r} = \frac{\Delta e_{l/r}}{\mathrm{d}t} R$, where  $R$ is the wheel radius, $\mathrm{d}t$ represents the time duration since the previous time step, and $\Delta e_{l/r}$ represents the average change in the wheel encoder tick value between the front and back wheel encoders since the previous time step. We use the following equation to calculate the vehicle's wheel velocities:
\begin{align}
    \dot{x}^w &= \frac{v_r + v_l}{2} \\
    \dot{\psi}^w &= \frac{v_r - v_l}{W}
\end{align}
where $W$ is wheel separation. The velocities are then integrated into a position measurement in the filter.

\subsubsection{Inertial Measurement Unit}

The IMU directly measures linear acceleration from the accelerometer, angular velocity from the gyroscope, and orientation from the magnetometer. We only fuse the 2D linear acceleration, yaw angular velocity, and yaw angle measurements ($\ddot{x}^a, \ddot{y}^a, \dot{\psi}^g, \psi^m$) according to our localized 2D environment assumption.

\subsubsection{GPR Relative Displacement}
We utilize a neural network, described in Section \ref{sec:lgpr-network}, to estimate displacement in the $x$-axis of the robot body frame. A set of $s$ consecutive traces $\mathcal{T} = \{\boldsymbol{t}_i, \dots, \boldsymbol{t}_{i+s} \}$, with each $\boldsymbol{t} \in \mathbb{R}^{200}$, are passed through the model weights to predict $\Delta d$, the linear displacement:
\begin{equation}
    \Delta d = \Phi(\mathcal{T})
\end{equation}
The number of traces used, $s$, is tuned as a hyperparameter alongside our network architecture hyperparameters. In order to integrate this into our measurement vector, we convert the relative displacement to an $(x, y)$ coordinate using the corresponding z-axis rotational angle $\psi_k^m$ from the IMU:
\begin{align}
    x_k^r &= x_{k-1}^r + \Delta d \cos{\psi_k^m} \\
    y_k^r &= y_{k-1}^r + \Delta d \sin{\psi_k^m}
\end{align}

We tune the covariance of the GPR network prediction to be higher while the robot is turning, as the displacement measured while rotating is dependent on where the GPR is mounted on the axis of rotation, which is a platform-specific setup parameter.

\begin{figure}[h]
    \includegraphics[width=0.7\linewidth]{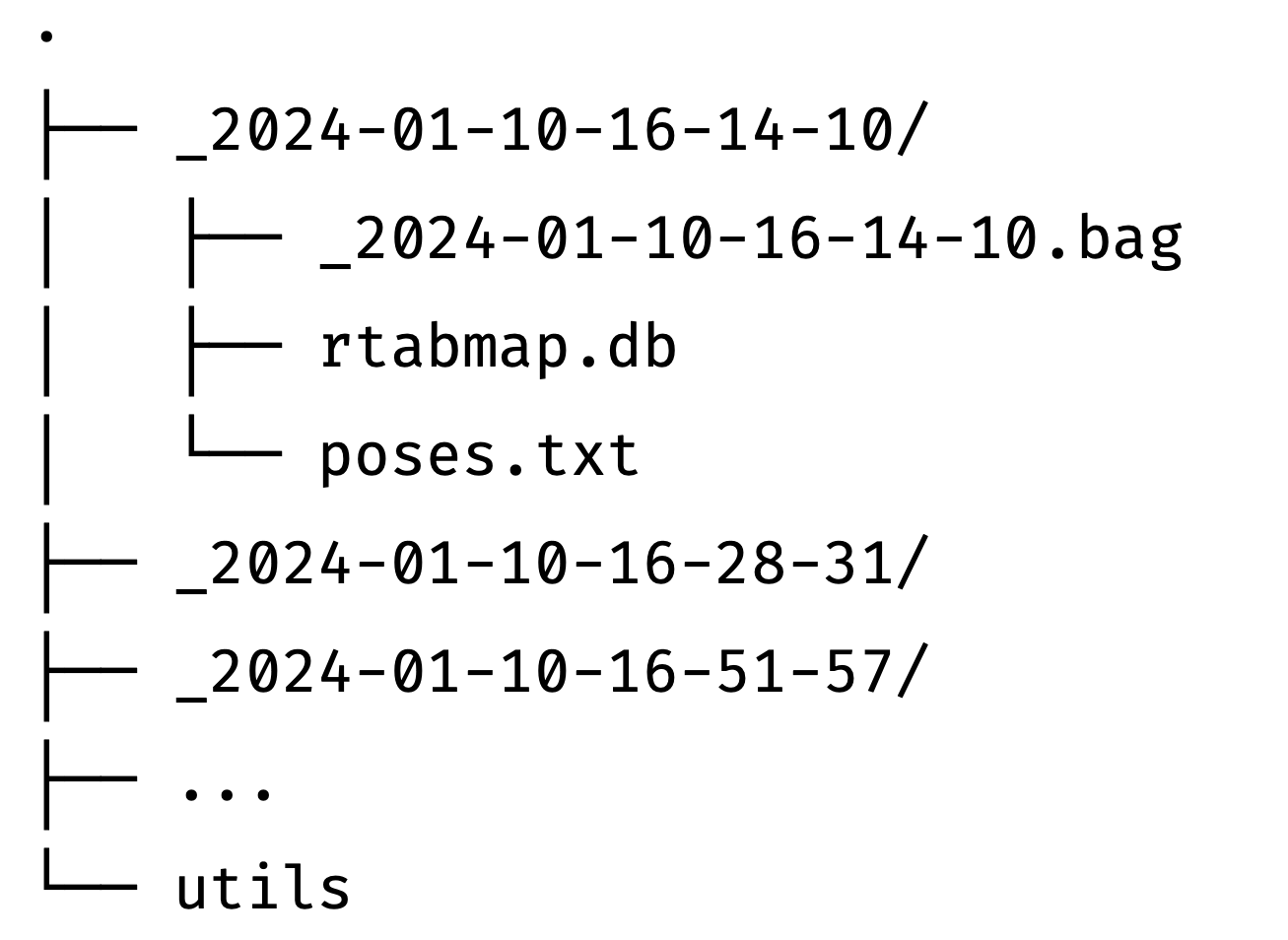}
    \caption{An overview of the dataset organization. There are 50 sequences, totaling around 1TB of data.}
    \label{fig:dir-struct}
\end{figure}

\begin{figure}[hb]
    \centering
    \includegraphics[width=\linewidth]{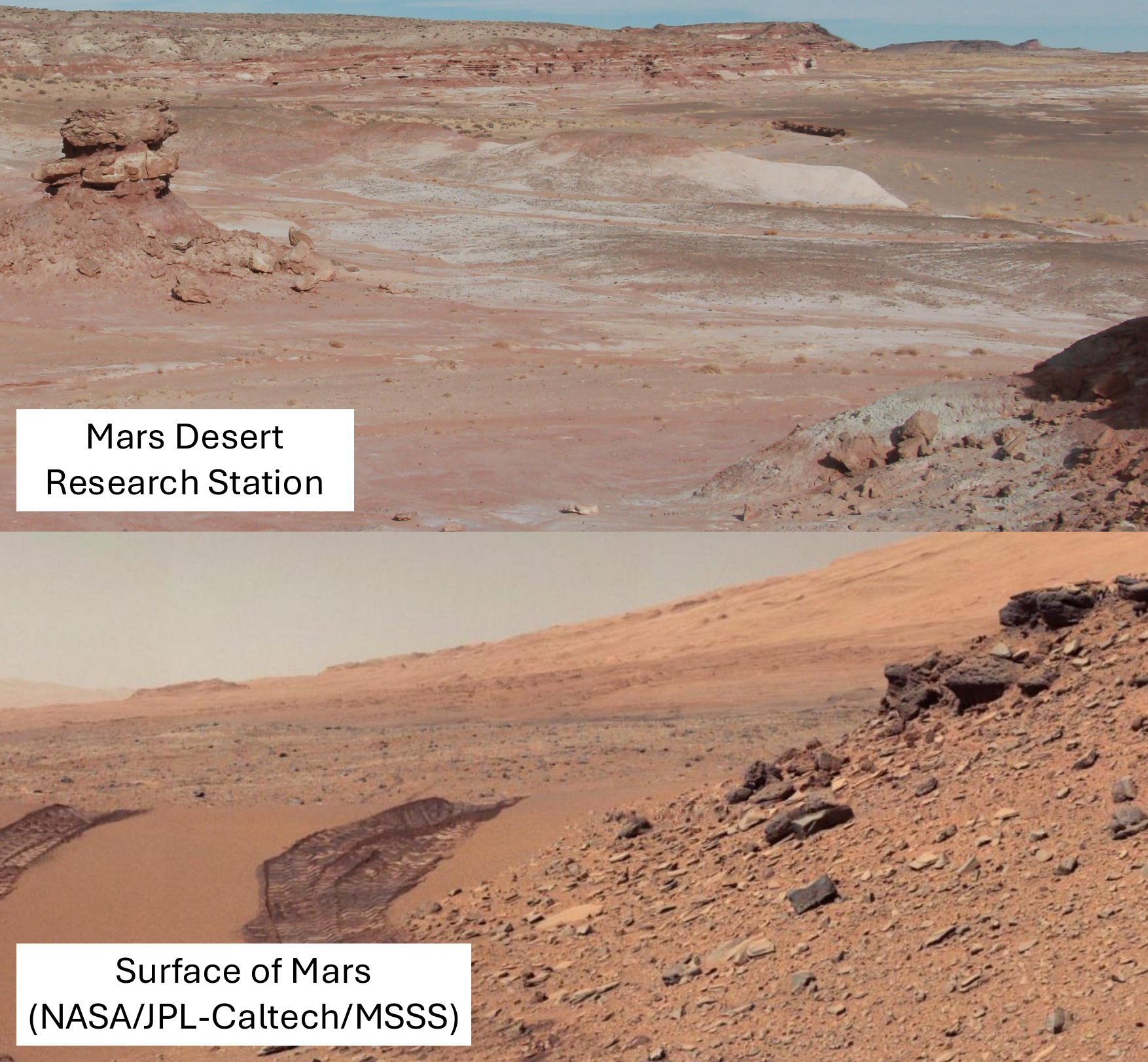}
    \vspace{-1em}
    \caption{Visual comparison of geologic features between the MDRS region (top) and Mars (bottom).}
    \label{fig:mdrs}
\end{figure}

\begin{figure*}
    \centering
    \includegraphics[width=\linewidth]{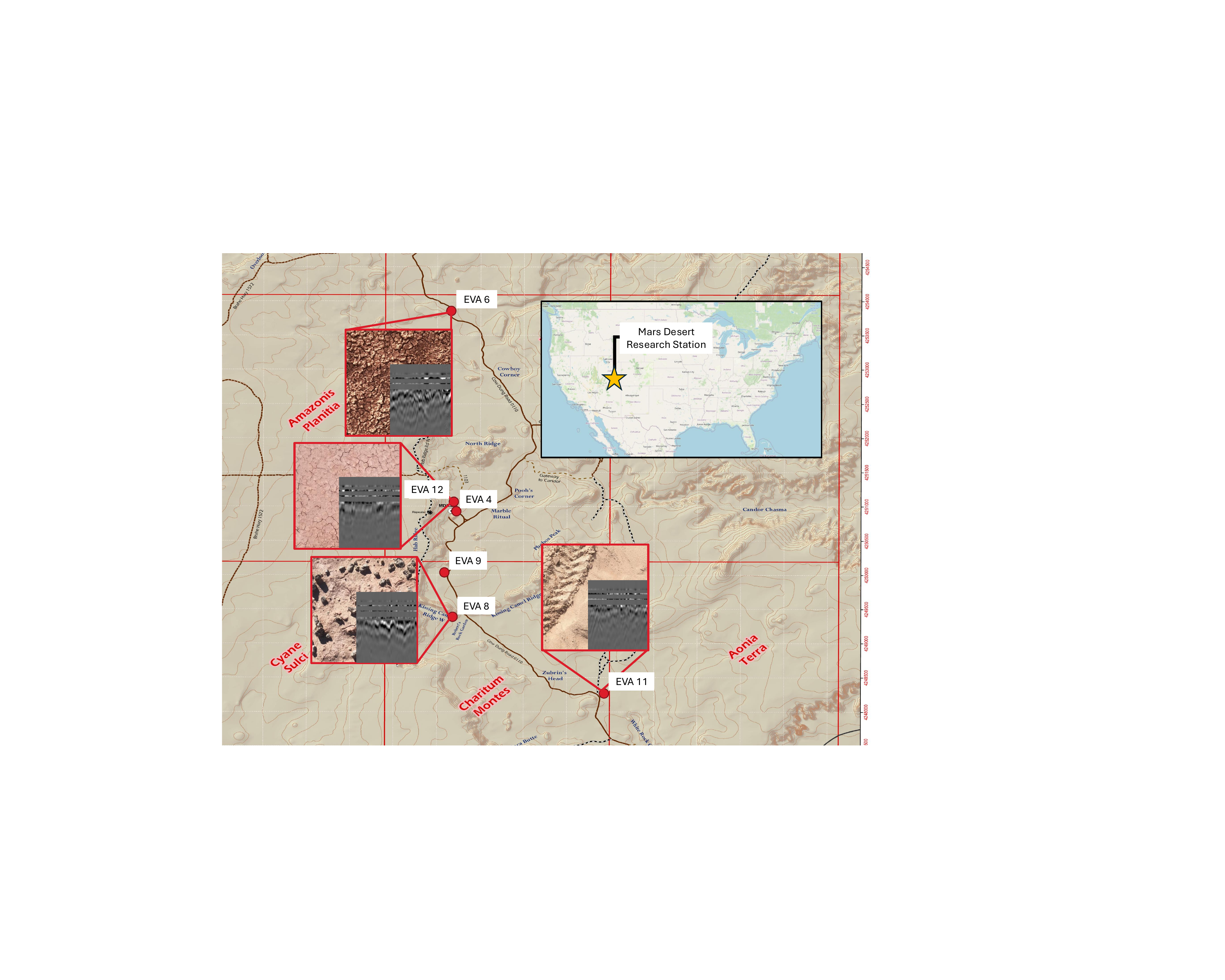}
    \caption{Areas of dataset collection in the region surrounding MDRS in southeast Utah. Note the variety of terrain types, shown in the red cut-outs, which provide a wide distribution of GPR signals for training the relative displacement network. A snapshot sample of a filtered GPR B-scan at each site is shown in the terrain image cutout. Base MDRS map provided by the Mars Society, and cutaway map of the United States provided by OpenStreetMap.}
    \label{fig:layout}
\end{figure*}

\begin{figure}
    \centering
    \includegraphics[width=0.9\linewidth]{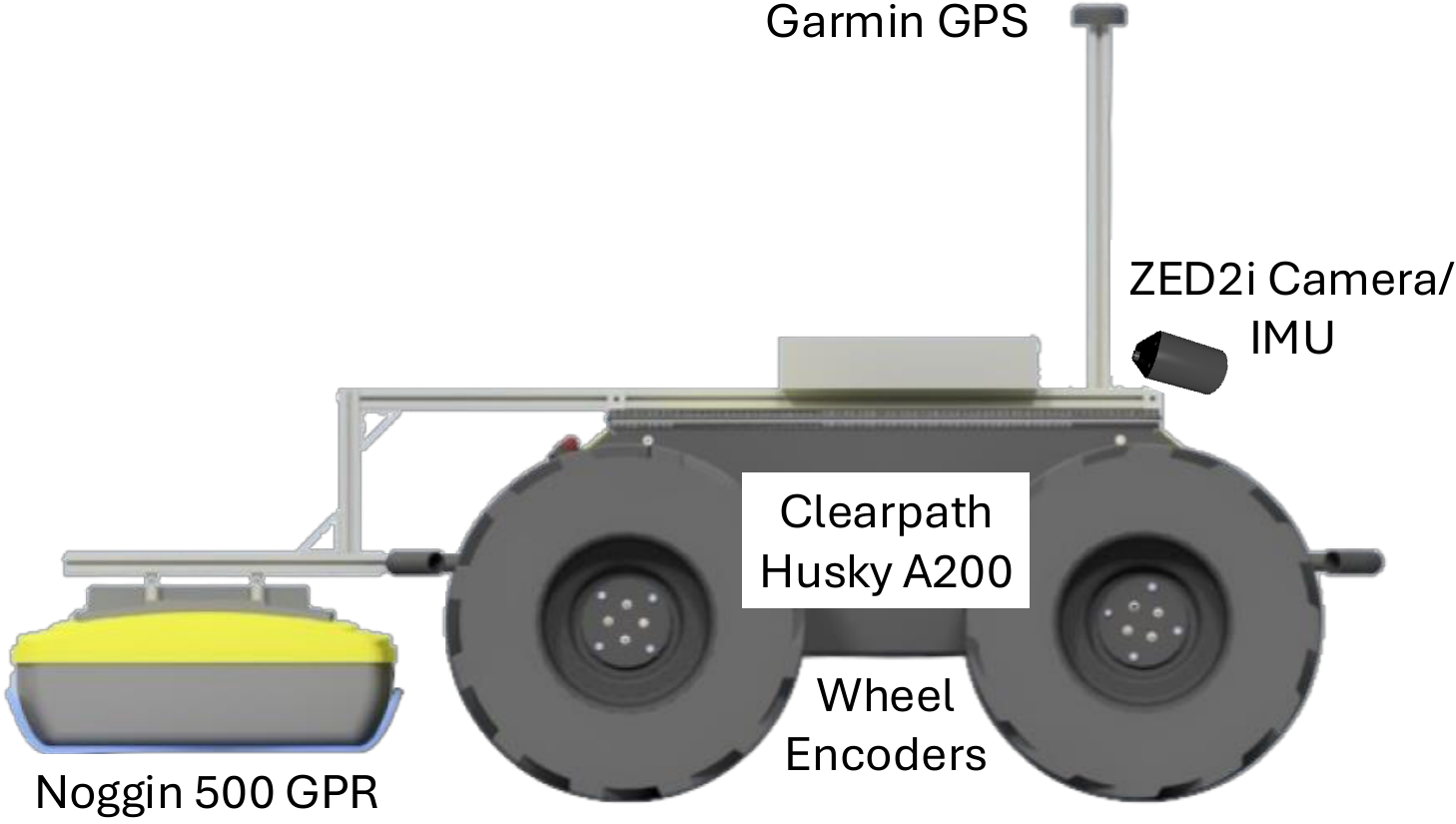}
    \caption{Diagram of custom Clearpath Husky A200 dataset collection platform, which was modified to include the NOGGIN 500 GPR sensor, a Garmin 18x module, and a ZED2i stereo camera.}
    \label{fig:husky}
    \vspace{-0.3em}
\end{figure}

\section{DATASET}

In order to evaluate a use-case for GPR-based localization for planetary rovers, it was necessary to collect a dataset at a site that is analog to the subterranean features that may be present on Mars. As previously outlined in Table \ref{tab:datasets}, the two existing LGPR datasets \cite{ort2023, baikovitz2021cmugpr} were collected on roads or in parking garages, which made them unsuitable for determining whether LGPR on Mars, the Moon, or rugged terrain is possible.

We collected a new dataset, MarsLGPR, at the Mars Desert Research Station (MDRS) near Hanksville, Utah over the course of 10 days in January 2024. The resulting dataset contains around 1TB of GPR, stereo camera, IMU, and robot wheel encoder data, which will be released with the directory structure shown in Figure \ref{fig:dir-struct}.

MDRS was chosen as a site for data collection due to its similarity to Martian geology (Figure~\ref{fig:mdrs}). A map of the collection sites is shown in Figure \ref{fig:layout}. The area is uniquely devoid of vegetation, and has evidence of similar geological processes as Mars, such as fluvial action, layered sediment deposits, presence of clays and salts, and sandy dunes \cite{stoker2011}.

The data collection platform was a Clearpath Husky A200, specially outfitted with a NOGGIN 500 GPR sensor. A ZED2i stereo camera with polarizer film was mounted on the front of the robot, which produced RGB images, depth images, and IMU data. The Husky wheel encoders track 78,000 ticks per meter. Wheel separation of the Clearpath Husky is $W = 0.165$ m, and the wheel radius is $R = 0.5455$ m. A custom ROS package was used to interface with the NOGGIN GPR sensor, which samples at a center radio frequency of 500 MHz. A diagram of the data collection platform is shown in Figure \ref{fig:husky}.

For each sequence, the vehicle was remotely controlled and driven around various terrain types, including sandy aeolian plains, cracked clays, and rocky washes. ZED stereo camera data was collected at 5 Hz and a resolution of 1280 x 720 pixels. GPR data was collected at 1.67 Hz in stacks of three repetitive samples to provide better control for noise.

\begin{table*}[!ht]
\centering
\caption{The cumulative displacement from the RTAB-Map (reference), Wick.*, wheel encoder, and GPRFormer network predictions using the RMSE metric in millimeters on the CMU-GPR dataset. Note that the CMU-GPR dataset was collected on flat concrete surfaces with little slip, making the wheel encoder very accurate. Bold is best.}
\label{table:rmse-comp}
\begin{tabular}{lcccccccccc}
\toprule
    && \multicolumn{4}{c}{CMU-GPR Dataset \cite{baikovitz2021cmugpr} (RMSE $\downarrow$)} && \multicolumn{4}{c}{MarsLGPR Dataset (RMSE $\downarrow$)}\\
    Method && Seq 1 & Seq 3 & Seq 7 & Seq 9 && 12-17-01-44 & 15-14-58-25 & 15-15-42-09 & 15-15-55-14 \\
    \midrule
    \midrule
    Wheel Encoder && 9.72 & \textbf{4.12} & \textbf{4.20} & \textbf{3.55} && 20.65 & 27.42 & 29.89 & 29.62 \\
    Wick.* \cite{wickramanayake2022} && 22.02 & 18.32 & 14.54 & 14.09 && - & 23.56 & 22.46 & 19.37 \\
    GPRFormer (Ours) && \textbf{8.62} & 7.64 & 6.51 & 5.98 && \textbf{15.59} & \textbf{13.32} & \textbf{16.92} & \textbf{17.58} \\
\bottomrule
\end{tabular}
\end{table*}

Note that we also collected GPS data on a Garmin 18x unit, however, due to an issue in parsing of satellite-based augmentation signals, the GPS data is not good enough to provide ground truth localization. Instead, we provide localizations from a map built and optimized offline by RTAB-Map \cite{labbe2013}, a popular VO and 3D reconstruction tool. VO is a well-posed task for the images in our dataset, as there is little camera motion blur and a large number of trackable features. These reference poses from VO are optimized offline, which is a computationally costly task and is infeasible to run in real-time on spaceflight hardware. However, we are able to use it as a reference trajectory for our model training and experiments in the absence of GPS.

In order to prepare the MarsLGPR dataset for training, we first remove anomalous traces where the recorded values are outside of the $\pm 50$ mV range and then we average the consecutive stacked traces to help reduce noise. Additionally, we perform linear and SLERP interpolation along the RTAB-Map poses to have time synchronized sensor and positional data, which is important for model training.

\section{EXPERIMENTS \& RESULTS}

In this section, we present experiments to validate our method qualitatively and quantitatively. We first present accuracy metrics on GPRFormer and two baselines: the wheel encoders and localization network Wickramanyake \cite{wickramanayake2022}. Then, we qualitatively and quantitatively evaluate the GPRFormer + EKF localization framework. Additionally, we include an ablation study and runtime metrics on GPRFormer. Experiments are performed on both the publicly available CMU-GPR dataset \cite{baikovitz2021cmugpr} and on our own MarsLGPR dataset.

\begin{figure*}
    \vspace{-1.5em}
    \centering
    \includegraphics[width=0.98\linewidth]{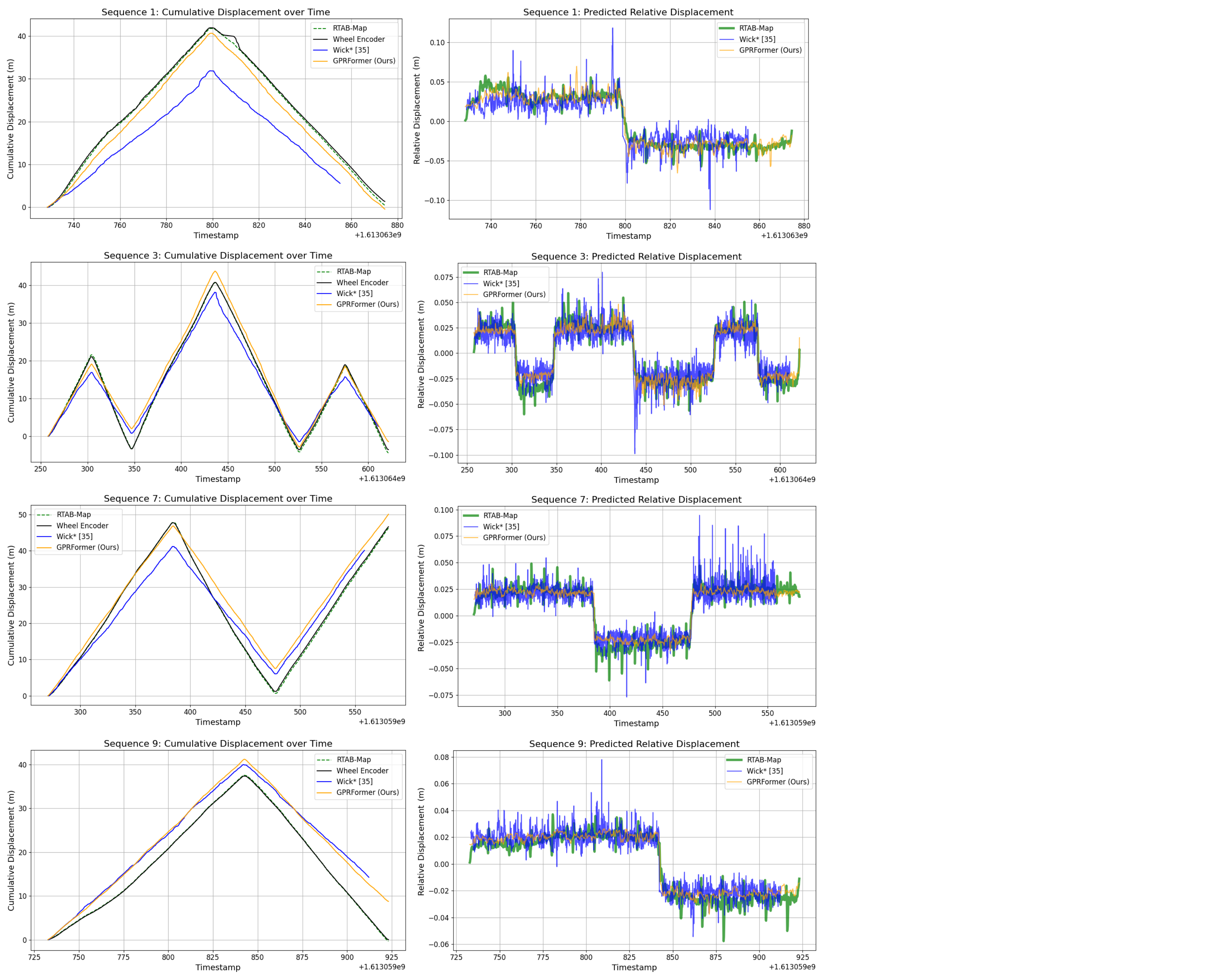}
    \caption{Comparison of GPRFormer relative displacement predictions against Wick.* \cite{wickramanayake2022}, the wheel encoders, and the ground truth on four test sequences from the CMU-GPR \cite{baikovitz2021cmugpr}  The left column shows the cumulative displacement over time, and the right column shows the accompanying predictions for the displacement at each time step.}
    \label{fig:qual_gprformer}
\end{figure*}

\begin{figure*}
    \centering
    \includegraphics[width=\linewidth]{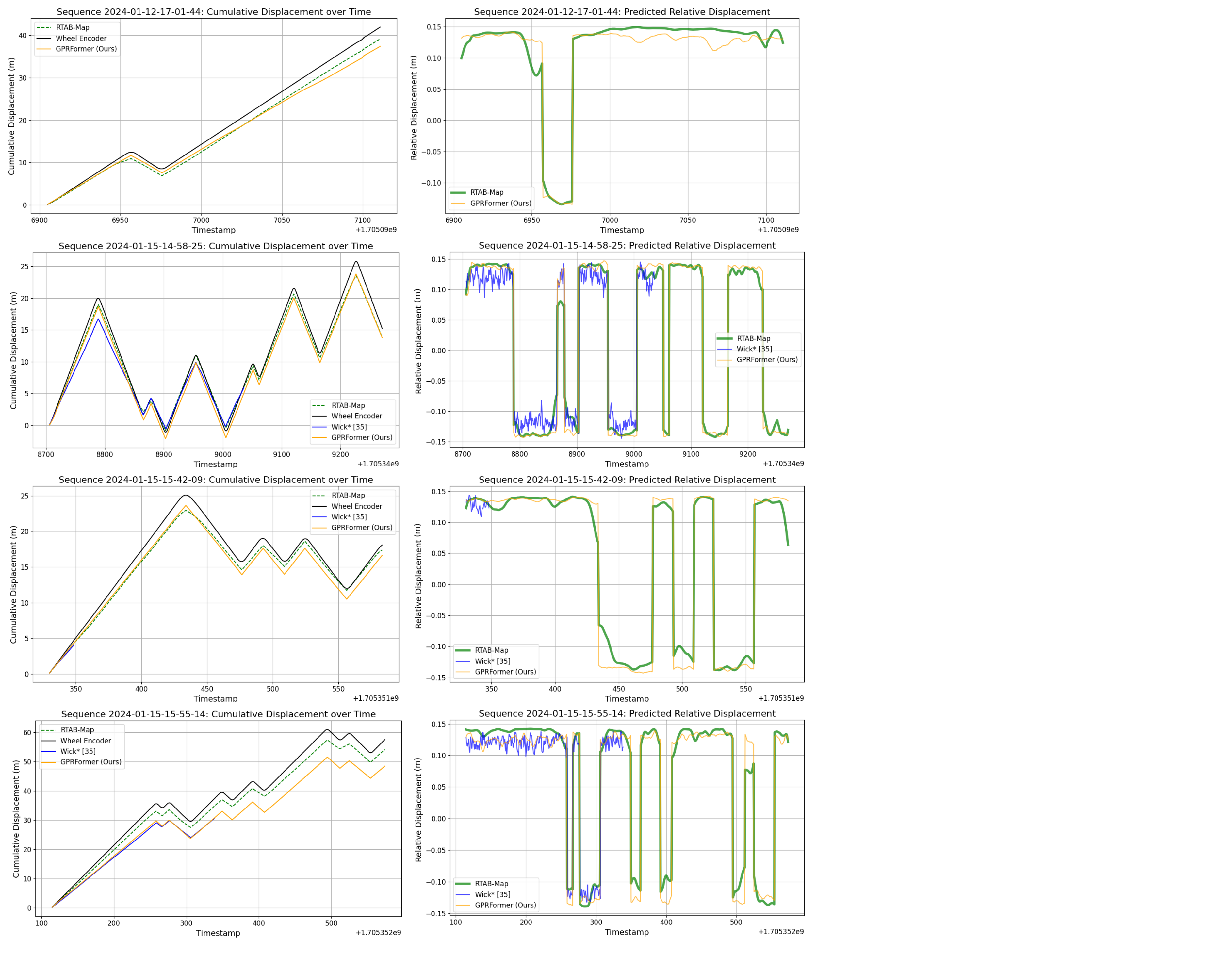}
    \caption{Comparison of GPRFormer relative displacement predictions against the wheel encoders Wick.* baseline, and RTAB-Map on four test sequences from the MarsLGPR dataset. The left column shows the cumulative displacement over time, and the right column shows the accompanying predictions for the displacement at each time step.  Note that the wheel encoders accumulate more displacement over time. This is due to the slippery terrain causing the wheels to spin out.}
    \label{fig:qual_gprformer_mdrs}
    \vspace{0.8em}
\end{figure*}

\subsection{GPRFormer Implementation Details}

We train two versions of our GPRFormer model, one on the CMU-GPR dataset and one on the MarsLGPR dataset. For the CMU-GPR dataset, we follow \cite{wickramanayake2022} and use sequences 0, 2, 4, 5, 6, 8, and 10 for training, and sequences 1, 3, 7, and 9 for testing. We apply background removal, dewow filtering, spreading and exponential compensation, and wavelet denoising to enhance feature visibility. During training, the model is supervised with displacement calculated from the ground truth RTK GPS positioning data for the CMU-GPR dataset and from reference RTAB-Map poses for the MarsLGPR dataset. The MarsLGPR train set contains 35 sequences and the test set contains four sequences.

The Adam optimizer is used with a batch size of 128, and a linear learning rate scheduler helps with better fine-tuning later in the training process. All training is performed on an NVIDIA GeForce 32 GB RTX 3090 GPU.

\begin{table*}[!htb]
\centering
\caption{Comparison of localization EKF with and without the GPRFormer relative displacement prediction. Trajectory accuracy is measured with RMSE ATE in meters, with a lower value indicating a closer trajectory to the ground truth. \textbf{Bold} is best.}
\label{table:rmseate-comp}
\begin{tabular}{lccccc}
\toprule
    && \multicolumn{2}{c}{CMU-GPR Dataset \cite{baikovitz2021} (RMSE ATE $\downarrow$)} & \multicolumn{2}{c}{MarsLGPR Dataset (RMSE ATE $\downarrow$)} \\
    Method && Sequence 3 & Sequence 7 & Sequence 01-12-17-01-44 & Sequence 01-15-14-58-25 \\
\midrule
\midrule
Encoder + IMU EKF && 3.010 & 2.113 & 4.947 & 8.485 \\
\midrule
Encoder + IMU + GPRFormer EKF && \textbf{2.903} & \textbf{2.100} & \textbf{2.576} & \textbf{8.136} \\
\bottomrule
\end{tabular}
\end{table*}

\subsection{Baseline Methods}
As no code is available for the Wickramanayake \cite{wickramanayake2022} baseline, we have re-implemented this method from scratch  in order to use it as a comparison. We differentiate our re-implemented version of this baseline as Wick.*. We directly follow the parameters provided in the paper, with a learning rate of 0.0001, an MSE loss function, the Adam optimizer, a batch size of 100, and the same train/test split. We trained the model for 250 epochs on the CMU-GPR dataset and 60 epochs on the MarsLGPR dataset, after which the MSE loss was no longer reduced. The paper reports batch MSE values of 0.266, 0.403, 0.115, and 0.143, our replicated version of the CNN-LSTM had batch MSE values of 0.370, 0.226, 0.123, 0.105 for Sequences 1, 3, 7, and 9, respectively.

\subsection{GPRFormer Accuracy}

We evaluate model performance using the Root Mean Squared Error (RMSE) of displacement estimates, reported in millimeters. RMSE is given by:
\begin{equation}
    \text{RMSE} = \sqrt{\frac{1}{T} \sum_{t=1}^T (y_t - \hat{y}_t)^2},
\end{equation}
where $y_t$ is the ground truth displacement, $\hat{y}_t$ is the predicted displacement, and $T$ is the prediction window. For GPRFormer, $T = k = 10$, and for Wick.*, $T = 2$. Since the two models generate predictions in overlapping windows of length $T$, we apply an overlap-add accumulation procedure to obtain a consistent per-timestep estimate: each window’s prediction is distributed evenly across its span and averaged with other overlapping contributions, producing a single displacement estimate per timestep. This binning ensures that different prediction granularities are made directly comparable, and also makes the predictions more robust to outlier results from the model.

Table \ref{table:rmse-comp} shows RMSE computed across the cumulative displacement for each test sequence in the CMU-GPR dataset and the MarsLGPR dataset. It is important to note that the wheel encoder data in the CMU-GPR dataset is very accurate, as demonstrated by the strong quantitative results. This is because the dataset collection occurred on concrete surfaces, where slip was minimal and there were no slopes. The Mars analog terrain in MarsLGPR proves more challenging for wheel encoder odometry estimation. GPRFormer demonstrates improvement over wheel encoder odometry for the more challenging MarsLGPR dataset. Our method, GPRFormer, outperforms the re-implemented Wick.* method on all eight test sequences across both datasets.



Figure \ref{fig:qual_gprformer} displays qualitative results from the RTAB-Map reference, the wheel encoder baseline, the Wick.* method baseline, and our model GPRFormer on the CMU-GPR dataset. The righthand column shows the individual GPRFormer predictions at each timestep compared with the displacement from the ground truth and wheel encoders. This allows us to inspect whether the network is able to learn local trends and patterns from the GPR data. The Wick.* predictions are very noisy, which is reflected in a higher RMSE metric in Table \ref{table:rmse-comp}. The lefthand column shows the cumulative displacement over the trajectory, with the direction of the wheel encoder (positive or negative) indicating the direction of movement. Note that in all eight plots, Wick.* has no results on the last section of the trajectory because the method relies on building up 202 consecutive GPR A-scans before making a displacement prediction.

\begin{figure*}
    \centering
    \includegraphics[width=\linewidth]{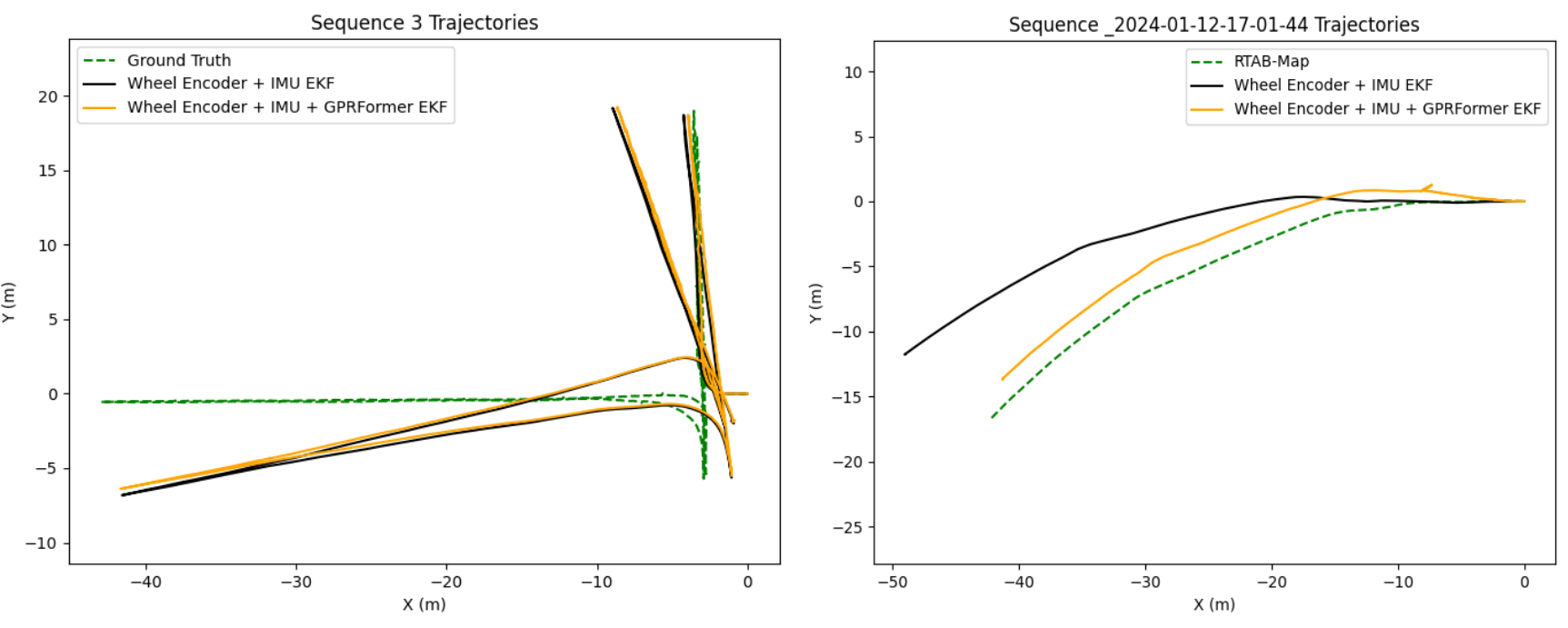}
    \caption{Qualitative comparison of filtered trajectories and ground truth on sequence 3 from the CMU-GPR dataset and sequence 12-17-01-44 from the MarsLGPR dataset. Note that the MarsLGPR dataset sequence shows much more improvement over the wheel encoder + IMU EKF due to the slippery terrain which hindered the wheel encoders.}
    \label{fig:qual_traj}
\end{figure*}

Figure \ref{fig:qual_gprformer_mdrs} shows qualitative results for GPRFormer and Wick.* on MarsLGPR test sequences. Note how the wheel encoder in the MarsLGPR sequences is much less accurate than the wheel encoder in the CMU-GPR dataset, which makes it a much less reliable source of odometry. Our GPRFormer network visibly outperforms the wheel encoder displacement as compared to the RTAB-Map reference, and this is also reflected quantitatively in Table \ref{table:rmse-comp}. We note that in Sequence 12-17-01-44, there were not enough GPR traces for the Wick.* method to make any predictions.

\subsection{GPRFormer Runtime Performance}

In this experiment, we discuss the ability of the GPRFormer model to operate in realtime. We present Table \ref{table:runtime}, which shows the runtime per sample in seconds for our method and baseline Wick.* \cite{wickramanayake2022}. We show runtime for inference on both GPU and CPU, as GPUs are typically inaccessible in space environments. Both methods have very fast runtimes, on the order of hundreds of Hz, although our method outperforms the Wick.* model.

Although the model runtimes are very fast, the most critical bottleneck for performance is the GPR data rate. For the MarsLGPR dataset, GPR data is collected at 1.67 Hz, meaning that our model (which requires 10 consecutive samples) could begin inference after 6 seconds, and then every half second after that. The Wickramanayake method requires two patches of 200 traces, with two traces in between, for a total of 202 traces (over 120 seconds of data collection) in order to begin inference.

\begin{table}[!htb]
\centering
\caption{Performance metrics (memory and runtime) for Wick.* and GPRFormer model inference.}
\label{table:runtime}
    \begin{tabular}{l|cccc}
    \toprule
    Model & \multicolumn{2}{c}{Avg. inference runtime (s)} & \multicolumn{2}{c}{Avg. memory (GB)} \\
    & CPU & GPU & CPU & GPU \\
    \midrule
    \midrule
    Wick.* \cite{wickramanayake2022} & 0.0706 & 0.0640 & \textbf{1.23} & 2.66 \\
    Ours & \textbf{0.0034} & \textbf{0.0032} & 1.90 & \textbf{1.39} \\
    \bottomrule
    \end{tabular}
\end{table}

\subsection{EKF State Estimation Accuracy}

We also demonstrate the resulting full state estimation results when integrating the GPRFormer predictions along with IMU yaw angle into an EKF to estimate $x$ and $y$ position. As GPRFormer is only able to predict 1D displacement, we increase the covariance of the GPR network predictions when the robot is rotating, as determined by the wheel encoder directions. Table \ref{table:rmseate-comp} shows a comparison of the EKF with and without the GPRFormer integrated in. We calculate the RMSE Absolute Trajectory Error (ATE) metric on both trajectories in relation to the ground truth after aligning them in the yaw axis.

For the CMU-GPR dataset, we note that the GPRFormer EKF only very slightly outperforms the wheel encoder baseline. This is because the wheel encoders are very accurate in this dataset, as the vehicle only drove over smooth covered surfaces. For the MarsLGPR dataset where there is confirmed wheel slippage on various rough terrains, we see a large improvement in the RMSE ATE. For one of the sequences, the error was cut in half after including GPR relative localization. The qualitative results in Figure \ref{fig:qual_traj} further confirm these results. We see that the inclusion of GPRFormer on the MarsLGPR sequence prevented the filter from overshooting the true trajectory due to wheel slippage.

The localization update rate is about 15 Hz, which is a satisfactory real-time rate for most applications.

\subsection{Ablation Studies}

In this experiment, we investigate how the different components of the network impact the overall model performance. We train five ablated versions of the model on the MarsLGPR dataset, with two additional experiments to further understand the $k$ and $\alpha$ parameters. See Table \ref{table:ablation} for a comparison of the performance of each ablation. The metric used in this experiment is RMSE in millimeters across all four MDRS test sequences, where lower is better.

\begin{table}[!htb]
\centering
\caption{Ablation study on GPRFormer network with results in RMSE (mm) across all four MarsLGPR test sequences. \textbf{Bold} is best.}
\label{table:ablation}
\setlength{\tabcolsep}{4pt}
    \begin{tabular}{ccccc|c}
    \toprule
        B-scan filter & encoder & dropout & seq pooling & \# layers & RMSE ($\downarrow$) \\
    \midrule
    \midrule
    \checkmark & \checkmark & \checkmark & \checkmark & \checkmark & \textbf{15.57} \\
     & \checkmark & \checkmark & \checkmark & \checkmark & 16.58 \\
     \checkmark & & \checkmark & \checkmark & \checkmark & 16.94 \\
     \checkmark & \checkmark & & \checkmark & \checkmark & 16.51 \\
     \checkmark & \checkmark & \checkmark & & \checkmark & 18.25 \\
     \checkmark & \checkmark & \checkmark & \checkmark & & 20.37 \\
    \bottomrule
    \end{tabular}
\end{table}

\subsubsection{B-scan filter pre-processing}
This pre-processing step helps make features in the B-scan more distinct and visible. We see in the ablation Table \ref{table:ablation} that not including this pre-processing step does make the results of the network worse, but not drastically so. This is likely because the network is still able to learn to find the finer-grained features in the B-scan even without the filtering. However, the inclusion of B-scan filtering likely makes the model learn more efficiently, resulting in an improvement when filtering is used.

\subsubsection{Linear encoder}
Similarly, removing the linear encoder from the GPRFormer architecture has a moderate effect on the overall network performance, slightly increasing the RMSE. This experiment demonstrates that the B-scan representation of the data is already quite compact and structured, which allows the model to learn well despite having no encoder. However, adding a linear encoder layer does allow for improved results by letting the model learn a more efficient latent space.

\subsubsection{Dropout}
Dropout is an important technique for improving network generalization and avoiding overfitting. In this variation, we reduce the dropout rate from 0.1 to 0 in order to understand the effects of this component in the architecture, which causes a moderate drop in network performance. The worse GPRFormer performance without the dropout module is likely caused by overfitting on the training set that occurs without this regularization.

\subsubsection{Dual sequence pooling}
As discussed in the technical approach of the paper, dual sequence pooling is a useful technique to allow the GPRFormer model to benefit from both the pre-transformer encoding and the post-transformer encoding by blending attention weights. In this experiment, we completely remove this from the network, which causes the network to strictly use the transformer encoder latent space. This greatly affects the performance of the model, with the average cumulative RMSE increasing to over 18 mm. This helps confirm that the inclusion of this attention weight pooling step, which balances between raw patterns and transformer-encoded patterns in the data, is a key component of the GPRFormer network.

\begin{figure}[!b]
\centering
\begin{tikzpicture}
\begin{axis}[
    title={Impact of $\alpha$ on model accuracy},
    xlabel={$\alpha$},
    ylabel={RMSE (mm)},
    height=5cm,
    width=0.95\linewidth
]
\addplot[
    color=green,
    mark=circle,
    ]
    coordinates {
    (0.1,16.9800)(0.2,16.8425)(0.3,16.6746)(0.4,17.5096)(0.5,16.4878)(0.6,16.5215)(0.7,15.5738)(0.8,17.2281)(0.9,19.7217)
    };
\addplot[
    color=green,
    mark=x,
    mark size=7pt,
    mark options={ultra thick},
    only marks,
    ]
    coordinates {
    (0.7,15.5738)
    };
\end{axis}
\end{tikzpicture}
\caption{A comparison of different $\alpha$ (dual sequence pooling weight) values and their impact on the model accuracy, as measured in RMSE. Note that a $\alpha = 0.7$ has the best results.}
\label{fig:a-plot}
\vspace{0.4em}
\end{figure}

Additionally, we include a further experiment on the optimal $\alpha$ value, which is the weighting parameter between the pre-transformer and post-transformer features. The $\alpha$ is a learned term, but for this experiment we fixed it to a static value in order to understand the impact of the term on the model performance. We see in Figure \ref{fig:a-plot} that when $\alpha$ is at a value of 0.7–which weights the post-transformer features a little higher than pre-transformer features–it has the best performance in terms of RMSE. When $\alpha$ is unfrozen during normal training, it converges to 0.70102, matching the results of this experiment.

\begin{figure}[!b]
\centering
\begin{tikzpicture}
\begin{axis}[
    title={Impact of $k$ on model accuracy},
    xlabel={$k$},
    ylabel={RMSE (mm)},
    height=5cm,
    width=0.95\linewidth
]
\addplot[
    color=blue,
    mark=circle,
    ]
    coordinates {
    (5,17.7853)(10,15.5738)(15,16.0005)(20,16.7262)(30,16.7607)(40,18.1925)
    };
\addplot[
    color=blue,
    mark=x,
    mark size=7pt,
    mark options={ultra thick},
    only marks,
    ]
    coordinates {
    (10,15.5738)
    };
\end{axis}
\end{tikzpicture}
\caption{A comparison of different $k$ (B-scan length) values and their impact on the model accuracy, as measured in RMSE. Note that a $k = 10$ has the best results.}
\label{fig:k-plot}
\end{figure}

\subsubsection{Multi-layer transformer encoder}
In this experiment, we drop the number of layers and attention heads to just one, and we see this has the most drastic effect on the network performance out of all of the ablations, with an increase in RMSE to over 20 mm. The transformer encoder is one of the core components of the network, and it has enough complexity that it can compensate for some of the other ablation experiments (removing the B-scan filtering step, for example). However, when this block in the network is simplified too far, it loses much of its ability to perform well.

\subsubsection{B-scan length k} 
In this experiment, we experimentally validate our choice to use a $k$ (B-scan length) of 10. In Figure \ref{fig:k-plot}, there is a clear minimum RMSE at $k = 10$, which is the value we used in our network design. As discussed in the paper sensor model, GPR has a very wide beam width, resulting in spatial overlap with adjacent A-scans. Stacking several traces together provides enough contextual information to the model about subsurface spatial features that the model is able to predict displacement. However, for the GPRFormer architecture the tradeoff between a larger B-scan and better performance starts to degrade after $k = 10$. The larger input makes it more difficult for the network to isolate meaningful features.

\section{CONCLUSION \& FUTURE WORK}

In this work, we proposed a novel framework to leverage GPR sensors aboard planetary rovers to aid in localization. Through our proposed network architecture, we demonstrated the ability for GPRFormer to learn relative displacement from GPR measurements on various types of Martian analog rugged terrain. We compare our method to both a wheel encoder and learned baseline \cite{wickramanayake2022}, and GPRFormer achieves superior results. Additionally, we integrate our prediction network into an EKF sensor fusion framework in order to provide efficient and accurate robot pose estimation. Our GPRFormer EKF outperforms the wheel encoder + IMU only filter, especially in areas of high wheel slippage. We will publicly release our MarsLGPR dataset and GPRFormer code to enable further work in GPR localization for Mars analog environments.

One limitation of the filtering-based framework that we use in this work is the need to carefully tune covariance matrices, which represent uncertainty from GPRFormer. Future work would include an investigation into quantifying the uncertainty from the GPRFormer model in order to provide dynamic and intelligent covariance matrices for uncertainty propagation in the filter. Additionally, we note that recent advances in 3D neural representations with novel sensors, such as NeRFs with Synthetic Aperture Radar \cite{lei2024} as well as 3D Gaussian Splatting with radar \cite{kung2025} and forward-looking sonar \cite{sethuraman2025}, motivate further work in GPR feature maps for localization.

\section*{ACKNOWLEDGMENT}
The authors would like to thank the mission support crew at MDRS and the Mars Society, namely Ben Stanley, for their support during field operations. Additionally, the authors would like to thank Madelyn MacRobbie, Rebecca McCallin, Ben Kazimer, Anna Tretiakova, and Wing Lam Chan for their assistance in conducting field data collection.

\bibliographystyle{ieeetr}
\bibliography{main}

\vfill\pagebreak

\end{document}